\documentclass{article}
\usepackage{ijcai21}

\usepackage{times}

\usepackage{soul}
\usepackage{url}
\usepackage[hidelinks]{hyperref}
\usepackage[utf8]{inputenc}
\usepackage[small]{caption}
\usepackage{graphicx}
\usepackage{amsmath}
\usepackage{booktabs}
\usepackage{adjustbox}
\usepackage{listings}
\title{Human-AI Symbiosis: A Survey of Current Approaches}
\author{Zahra Zahedi
\and
Subbarao Kambhampati
\affiliations
CIDSE, Arizona State University\\
\emails
\{zzahedi, rao\}@asu.edu,

}
\begin{document}

\maketitle

\begin{abstract}
In this paper, we aim at providing a comprehensive outline
of the different threads of work in human-AI collaboration. By highlighting various aspects of works on the human-AI team such as the flow of complementing, task horizon, model representation, knowledge level, and teaming goal, we make a taxonomy of recent works according to these dimensions. We hope that the survey will provide a more clear connection between the works in the human-AI team and guidance to new researchers in this area.
\end{abstract}
%\vspace{-10pt} 
\section{Introduction}
As AI systems are becoming a part of our day-to-day lives, there would be more interest to have a team of humans and AI where they can improve the shortcomings and limitations of one another if they were otherwise working alone. Even though when humans and AI team up, they can empower each other's abilities and reach better outcomes, new challenges emerge that aren't there with AI systems or humans alone. The primary challenge of an AI agent that is functioning alone is how effectively and flawlessly it achieves its goal. However, in a team of humans and AI assisting each other for a team goal, the challenges are not limited to the goal itself, but the AI system should also have the ability to not only reason about the human's actions but their mental models.\\
There exists a variety of works in this area that try to have human and AI work together as a team for better outcomes. While the works in this area focus on different challenges, the lack of coherency between them makes it hard to see a clear connection between these works. In this paper, we try to categorize research efforts into different dimensions to have a taxonomy of related works in this area. To this end, we introduce various aspects of works in human-AI teams, then we elaborate on how the recent works in this area are matched with those aspects. This effort can hopefully provide a clear connection between researches in human-AI team from different perspectives and guidance for future research.\\
\noindent\textbf{Survey Scope and Outline} In this survey, we highlight how different works in the area of human-AI teams can be viewed and organized from different dimensions. First, we emphasize how how does the complementing flow between the human and the AI, then investigate task horizon and model representation in different works. Also, we organize different works in this area based on their knowledge and capability levels and their teaming goal perspectives. Then, we highlight how recent works can be categorized regarding these dimensions.
%\vspace{-8pt} 
\section{Dimensions of Human AI Teams}
\begin{figure*}[t]
    \centering
    \includegraphics[width=\textwidth, trim=0 232 0 0, clip]{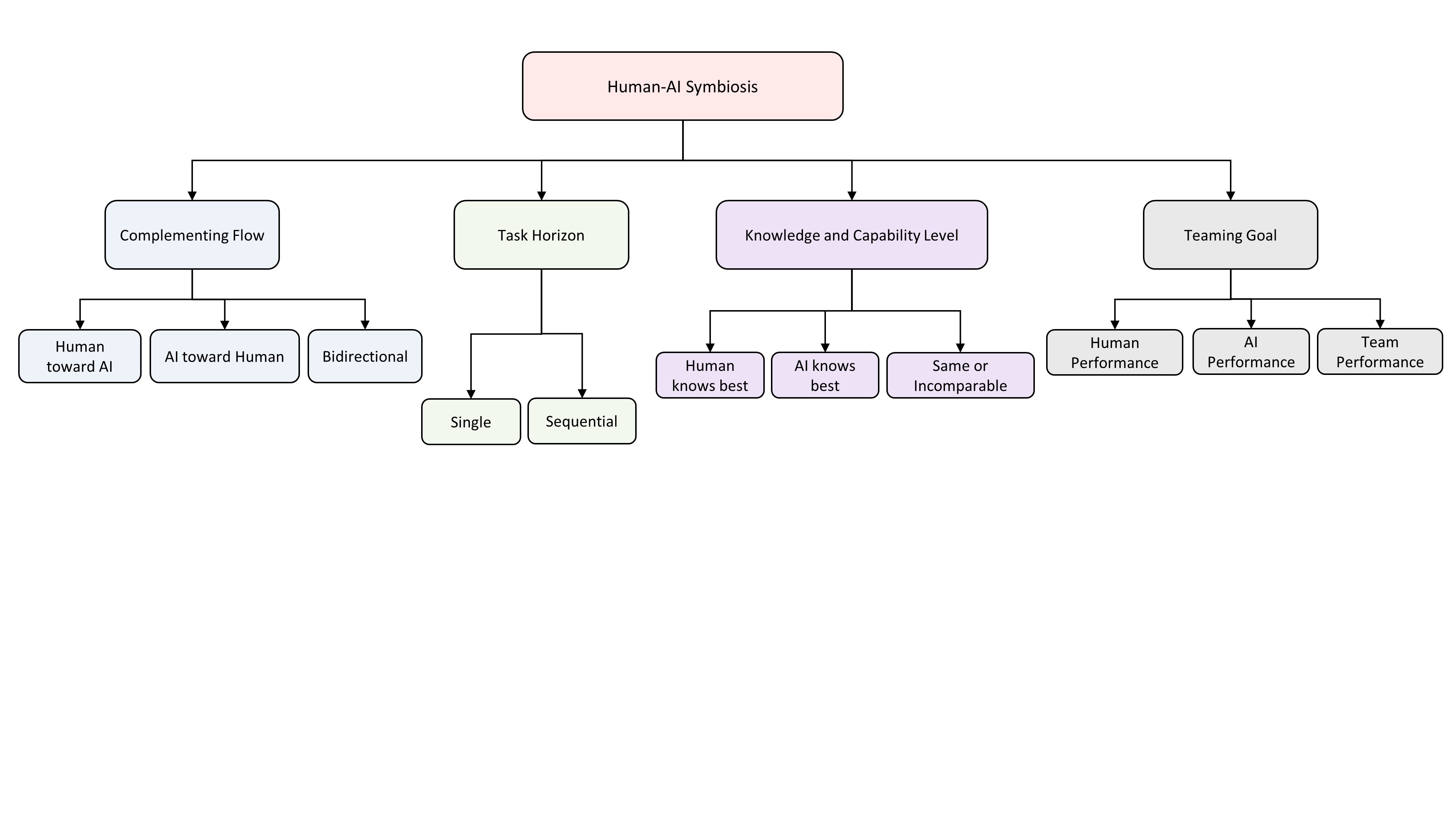}
    %%\vspace{-5pt} 
    \caption{We survey various dimensions based on the paradigm of recent approaches}
     %\vspace{-8pt} 
    \label{fig1}
  
\end{figure*}

In this section, we highlight many dimensions upon which the recent works in the area of human-AI teams can be integrated. In particular, we focus on different dimensions that make the works in this area distinct from each other (shown in Figure \ref{fig1}). Thus, we first provide an overview of the different dimensions associated with this direction of research, then various aspects of them delve into a survey of the existing works.
 %\vspace{-6pt}
 
\subsection{Complementing Flow}
When we have a human-AI team, it is really important to see who is complementing whom. Thus, in the team in which they try to compensate for their weaknesses, it might be the human who complements the AI agent or the AI agent who complements the human, or even it can be considered as a peer to peer complementing where both entities are complementing each other. Different challenges arise depending on which category is researched. Therefore, (1) when the human complements the AI system, human inputs should be used to improve the AI system's performance in which case in addition to the challenges of using human inputs to complement, which itself is associated with its own costs, constraints, and quality, and availability issues, the AI system also needs to have some kind of reasoning capability to know how and when to use the human inputs \cite{kamar2016directions}. However, (2) when the AI is complementing the human, not only the AI should help, it's important that the human also recognizes the help. So, since it is really important that the human understands the AI, in addition to the challenges that stem from the task itself and the optimality and effectiveness of the outcome, the interpretability of the AI behavior plays an important role. (3) For a peer-to-peer teaming, the AI agent is still helping the human, but the bidirectional communication and feedback between the two entities will help achieve a more effective teaming.
 %\vspace{-6pt}
\subsection{Task Horizon and Model Representation}
\textbf{Task Horizon} is another aspect that separates the works in this area. Generally, the tasks can be categorized as (1) single tasks such as classification and prediction or (2) sequential tasks which are sequential decision-making problems such as planning and scheduling. \\
In one category, we have AI systems that can complement humans for perceptual, diagnostic, and reasoning tasks. These AI systems are usually Machine learning (ML) models trained to complement the strength of the human for predicting the answer to a given task \cite{bansal2019beyond}. Examples of these predictive models are a medical decision support system used by a doctor \cite{wilder2020learning,bansal2019beyond}, a recidivism predictor that advises a judge \cite{wilder2020learning,tan2018investigating,lakkaraju2019faithful}, or a classification system that helps scientists to understand the distribution of galaxies and their evolution \cite{wilder2020learning,kamar2016directions}.\\
On the other hand, we have tasks that are sequential decision-making problems. This can be a proactive decision-making system that helps the user in constructing a plan \cite{sengupta2017radar,sengupta2018ma}, a robot which is making a plan to help the human in doing a task \cite{unhelkar2018human,dragan2013legibility,chakraborti2019explicability}, or a scheduling system that helps to allocate multiple users to different tasks \cite{manikonda2014ai,zahedi2020not}.\\
\noindent\textbf{Model Representation} is another variation in this area. While any background knowledge of the human or the AI agent is considered as the model, since it is hardly possible to have a model that captures all the background knowledge, the abstraction of this is shown in different forms. Thus, this abstraction of the model can be the beliefs or state information of the agent such as its goals and intentions, its capabilities or initial conditions or the reward function \cite{chakraborti2019explicability}, the error model \cite{bansal2019beyond}, features, and decision logic rules of a classifier \cite{lakkaraju2019faithful}. It can also include the observation model and the computational capability of the observer \cite{kulkarni2019unified}.
%\vspace{-7.5pt} 
\subsubsection{Relationship between Tasks and Models}
Although for the works mentioned in this survey, we investigate the task horizon and the model representation separately, there is a clear relationship between them. Since the works with the single task such as classification, prediction, etc. are using machine learning models, they usually consider the model as error function, features and probability distribution over output data. However, in sequential tasks, we usually have Markov Decision Process (MDP) models along with other factors such as initial conditions, goals, and observation models. Moreover, the new direction of works in this area considers human trust as another element of the model \cite{chen2018planning,xu2015optimo}.
 %\vspace{-6pt}
 
\subsection{Knowledge and Capability Level}
A valid reason for having human-AI teams is to achieve a complementing performance that is better than either one of them on their own. However, this is only possible if the appropriate tools are leveraged. The capability level and knowledge of either is very important to achieve a real complementary performance. For instance, in the case of having the AI agent assisting the human, appropriate reliance is crucial to improve performance in a team \cite{bansal2020does}. So, over-reliance on a human with limited computational capability or an AI agent with limited knowledge not only cannot improve team performance but can hurt it. The literature in this area considers various scenarios regarding the capability and knowledge levels of the AI agent and the human, and they base their work on them. Therefore, the level of knowledge and capabilities of the AI agent compared to the human can affect the types of problems that can be solved. Works in this area are usually categorized into (1) The AI agent knows more and so its model is the right one (2) The human knows more (3) Both have the same capability and knowledge level and (4) Their knowledge and capability level are not comparable or are unspecified.
 %\vspace{-6pt}
\subsection{Teaming Goal}
A human-AI team can face different challenges depending on its goal. Whether the purpose of the team is to improve the overall performance, the human performance, or that of the AI agent will result in different challenges.
Another aspect is the interaction state of the team as the goal of the human-AI team can be affected differently with single interaction versus multiple interactions between the human and the AI.
\begin{table*}[tp]
\centering
\caption{Summary of results considering complementing flow, task horizon, and knowledge and capability level. Note for reviewers: We had to go with the \textsf{plain} bib format so we can tabulate the paper numbers}
\label{tab}
\begin{adjustbox}{width=1\textwidth}

\begin{tabular}{l|p{22.1mm}p{22.1mm}p{22.1mm}p{22.1mm}p{22.1mm}p{22.1mm}}
\toprule
 & \multicolumn{6}{c}{Knowledge and Capability level}\\
\cmidrule{4-5}
Complementing Flow&\multicolumn{2}{c}{Human knows best}&\multicolumn{2}{c}{AI knows best} &\multicolumn{2}{c}{Same or Incomparable} \\
\cmidrule{2-7}
 & Single Task & Sequential Task & Single Task & Sequential Task & Single Task & Sequential Task \\
\cmidrule{1-7}

Human Complement AI & \cite{kamar2015planning,kamar2013lifelong,kamar2013light,wilder2020learning,nushi2017human}  &\cite{gao2015acquiring,torrey2013teaching,amir2016interactive,talamadupula2017architectural,kamar2015planning,kamar2013lifelong,kamar2013light,wilder2020learning}  &  & &\cite{tan2018investigating} &\cite{ramakrishnan2019overcoming} \\
AI Complement Human &  & &\cite{bansal2019updates,bansal2020optimizing,lakkaraju2019faithful,nushi2018towards} &\cite{zhang2017plan,kulkarni2019explicable,dragan2013generating,dragan2013legibility,dragan2015effects,macnally2018action,kulkarni2019signaling,kulkarni2019unified,kwon2018expressing,chakraborti2017plan,sreedharan2017explanations,raman2013sorry,raman2013towards,briggs2015sorry,eriksson2017unsolvability,eriksson2018proof,chakraborti2017balancing,sreedharan2019expectation,sreedharan2019can}& &  \\
Bidirectional Complementing &  &  & &\cite{unhelkar2018human,grover2020model,sreedharan2018hierarchical,zahedi2020not,valmeekam2020radar,chakraborti2015planning,zhang2012human,manikonda2014ai,sengupta2017radar,sengupta2018ma,grover2020radar} & &\cite{unhelkar2018human,ramakrishnan2019overcoming,huang2019nonverbal} \\

\bottomrule
\end{tabular}

\end{adjustbox}
%\vspace{-0.2cm}
\end{table*}
 %\vspace{-6pt}
\subsection{Scope of the Survey}
Integrating the abilities of humans and AI in a team offers great promise for the development of practical applications. This is a growing field of research with many challenges. The recent works in this area try to address many of the existing challenges, but there are significant differences between the direction of the works in this area such that this makes them seem independent and separated. The four dimensions we proposed in the paper can be a standpoint to see a clear connection as well as the differences between the existing works.  While all four mentioned dimensions are significantly important and discriminative, the first dimension alone can bring up distinct challenges for each category. Thus, we describe the recent works through the challenges that arise with the category of the first dimension as the central one, then elaborate other dimensions through them to make a cluster of mutual works. Table \ref{tab} is the summary of results from three dimensions.
% \section{Complementing Flow}
% With the advances in AI systems, it is expected that AI systems come naturally so that they act accurately to achieve the goal and in a way that makes sense to the world and humans. However, if the AI systems function alone, it is hardly possible to have systems that are understandable and accurate. So, they may make mistakes and fail from time to time or act such that the actions don't make much sense. Moreover, as the tasks become more complex and vital, mistakes can result in irreversible consequences and losing users' trust. Thus, instead of having AI systems that function alone, having the human in the loop of AI systems not only can complement the AI capabilities to perform better but also human involvement can prevent AI failures and shortcomings, and can help for having more interpretable systems.\\
% In this 
%\vspace{-8pt} 
\section{Human Complements AI}
It is acknowledged that to overcome AI mistakes and limitations the human involvement is necessary. However, using human inputs to improve the AI systems' performance has many challenges. These challenges include factors associated with costs, constraints, quality, and availability when using human inputs to complement, and the need for the AI system to have some kind of reasoning capability to know how and when to use the human inputs \cite{kamar2016directions}. In this section, we discuss the different ways that the human can complement an AI system and the existing challenges and we investigate other introduced dimensions through them.
 %\vspace{-6pt}
\subsection{
%Infusing human intelligence to 
Solving the Task}
One way to complement an AI with human input is to infuse human intelligence (e.g. inputs from crowd-sourced workers) to improve the accuracy of the given task. In such a setting, the challenges would be reasoning about when and where those inputs can be used to reach a better efficacy. Since the human complements the AI system, the works in this line mostly assume that the human (which are crowdsourced workers) has more knowledge and capability in solving the task. Moreover, the teaming goal in this area is how to improve the AI performance while optimizing the cost of collecting information from the human, which result in improving the efficacy of the large-scale crowdsource. Thus, the task horizon is separated into two parts (1) the learning part which is the prediction of the answer, and (2) the inference part for reasoning and planning about the hiring and routing of workers. With this, the task horizon for the first part is a single task classification task given the input data, however, for the second part, it is a sequential decision-making problem using different planning methods. For example, the Crowdsynth effort describes a general system that combines machine learning and decision-theoretic planning to guide the allocation of human efforts in consensus tasks \cite{kamar2015planning}. By collecting multiple assessments from human workers, their goal is to identify the true answer to each task such that the AI agents learn about the task and capability of the workers to make decisions about how to guide and fuse different contributions.\\
Furthermore, they extend Crowdsynth for solving hierarchical consensus tasks (HCT) to find a true answer to a hierarchy of subtasks \cite{kamar2013light}. They described a general system that uses hierarchical classification to combine evidence from humans in various subtasks with machine perception for predicting the correct answer. They used Monte Carlo planning to reason about the cumulative value of workers for the decision on hiring a worker, and customized it for HCT to constraint the policy space. \\
CrowdExplorer is another extension for the adaptive control of consensus tasks when an accurate model of the world is not available and needs to be learned \cite{kamar2013lifelong}. CrowdExplorer is using a set of linear predictive models and a novel Monte Carlo planning algorithm to continuously learn about the dynamics of the world and simultaneously optimize decisions about hiring workers and reasoning about the uncertainty over models and task progress in a life-long learning setting.\\
Moreover, other than consensus tasks, one of the important challenges is the ability to make a balance between value and costs of collecting information prior to taking an action. This is the reasoning behind whether to stop or continue collecting information (human inputs). Since the individual observation is weak evidence, the computation of the value of information where there is a large sequence of evidence is challenging. Monte Carlo value of information (MC-VOI) performs a large look-ahead to explore multiple observation and action sequence with a single sample \cite{kamar2013light}.\\
Furthermore, unlike the standard approaches which construct a machine learning model to predict the answer to a given task and take the predictive model as fixed and then build a policy for deciding when to use human inputs, the authors in \cite{wilder2020learning} jointly optimize the predictive model and query policy with a combined loss function that puts into account the relative strength of the human and machine.\\
Although in the aforementioned works, the model is represented through features, the value of information, and the probability distribution over different answers to the tasks, there are researches in which the crowd-sourced inputs help the planner to build a domain of the model which includes state information, goal and initial state to solve a planning problem \cite{gao2015acquiring}. In such works, the challenge is how to exploit the knowledge to address the noisy inputs from the crowds.
 %\vspace{-6pt}
\subsection{Troubleshooting}
To reach a better competency, AI systems should be able to identify and troubleshoot their failures. Using human inputs can help to effectively identify the failures and try to address them accordingly. This can be done either (1) by investigating and identifying the differences between the human and the AI agent in doing the task when both the human and the AI agent may have their own shortcomings. This means the knowledge and capability level of both the human and the AI agent are the same or incomparable, or (2) by getting feedback and assessments from humans to know and address the failures when the human is the expert which categorizes as the human having a higher knowledge and capability level. For instance, the effort on analyzing how human and machine decisions differ and how they make errors on the problem of Recidivism prediction may yield improvements in the Recidivism prediction \cite{tan2018investigating}. So, they used a widely used commercial risk assessment system for the Recidivism - COMPAS, and characterized the agreement and disagreement between the human and the COMPAS by clustering and decision trees, then investigate how combining the differences can reduce the failures. 
The systematic errors result from the difference between the simulated world and the real world -blind spots- can be addressed by human inputs, because the agent may never encounter some aspects of the real world \cite{ramakrishnan2019overcoming}. In this work, they applied imitation learning to demonstrate data from the human to identify important features that the human is using but the agent is missing and then they used the noisy labels extracted from action mismatches between the agent and the human across simulation and demonstration data to train blind spot models. Regarding the knowledge and capability level and teaming goal, both of these works have an incomparable level of expertise between the human and the AI, and both try to improve team performance as the teaming goal. However, their model representation and task horizons are completely different. \cite{tan2018investigating} represents the model as the error with a single task horizon that is prediction, and \cite{ramakrishnan2019overcoming} has sequential decision-making tasks with features, actions, and rewards represented as the model.\\
Using human intellect, when the human is considered more expert, assessing the system can result in the troubleshooting of the system failures with the goal of improving AI system's performance. The effort by \cite{nushi2017human}, simulates potential component fixes through human computation tasks and measures the expected improvements in the system. The system is first evaluated by crowd-sourced workers then when the workers apply their fixes for the components, the fixed output is integrated into the system, and the improved system is evaluated again by crowd-workers so that the fixes of earlier components are reflected on the inputs of later components.
 %\vspace{-6pt}
\subsection{Acting in Unknown Environments}
One of the characteristics of AI systems that would act naturally is their ability to deal with new environments and tasks. For agents to act in new environments, one way is to learn how to act in such environments, so having human inputs like advice or instruction would help significantly for more effective learning. Thus, in such works, the human is considered as an expert with higher knowledge and capability level, and the teaming goal is to improve the AI agent performance. Therefore, the human can be like a teacher that gives instructions to the AI system (student) by suggesting actions that the AI agent can take while learning \cite{torrey2013teaching}. The authors proposed a different set of teaching algorithms such as early advising, importance advising, mistake correction, and predictive advising that a human teacher can take to show how they affect the learning speed of an RL agent which is learning how to act. However, their proposed method required the human to continuously monitor the AI agent to know when and where to give advice, so an interactive teaching strategy in which the teacher and the student jointly identify the advising opportunities will address this issue \cite{amir2016interactive}. When the human teacher and AI student interactively train such that the RL agent decides when to ask for attention and the human teacher who is asked for the attention decides what advice to give, it can speed up AI agent learning without the need for constant attention. \\
Moreover, for the AI agent to use human instructions to understand the different aspects of an unknown environment like tasks, goals, subtasks and other unknowns, needs a mechanism for understanding human instructions in natural language \cite{talamadupula2017architectural}.
%\vspace{-8pt} 
\section{AI Complements Human}
With the advances in Artificial Intelligence, there is the pervasive use of AI systems to integrate their capabilities with human users. As a result of ubiquitous AI systems which are helping humans in their tasks, the first challenge is regarding the optimality and efficacy of AI systems in doing different tasks and decision making to achieve the desired complementing objectives. However, unlike the works in which the human complements the AI where there was significant attention to the helper's costs and constraints, here the focus shifts more to the understandability of the help. Indeed, the help is considered efficient if AI systems behavior conforms to the human's expectation, and human trust. Although there are an ever-expanding line of works that are investigating the AI agent physical and algorithmic capabilities so that they will be able to participate in a variety of complementing tasks and interactions autonomously \cite{scheutz2007first}, in this section, we just talk about different aspects that AI systems can take into account for being interpretable and trustable to be an effective complement toward the human along with discussing how works in this area are different in regard to the introduced dimensions.\\
The main challenge in having more interpretable and trustable systems is how to account for the human mental model. A behavior might be uninterpretable to the human if it's not comprehensible with respect to the human's expectations \cite{chakraborti2019explicability}. This mental model like the AI model can be regarded as the beliefs, state information, goals, intentions, capabilities, reward function, features, and errors, but it might be different from the AI model. For instance, when the human interacts with an AI agent, they make a mental model of the AI agent's error boundary which affects the human's decision as to decide when and where to trust and use the AI agent's complement \cite{bansal2019beyond}. Thus, the AI agent can account for this mental model to optimize for team performance instead of mere accuracy \cite{bansal2020optimizing,bansal2019updates}.\\ Therefore, given the human mental model, the AI agent can behave according to the human's expectations or communicate to change the expectations. The teaming goal for the works in this area is to improve team performance. Although in most of the works the AI agent is the only actor in the team, interpretability will affect the team performance in longitudinal interactions. Moreover, while most of the works in this area try to improve the team performance, and the AI agent is considered to have more capability and knowledge than the human, their task horizon and the model representation are different depending on various interpretable communications.
 %\vspace{-6pt}
\subsection{Interpretable Through Behavior}
The interpretable behavior can be concerned with the plan or the goal. The AI agent in a sequential decision-making task horizon can behave to be understandable and predictable to the human by showing an explicable plan or predictable plan. Where explicability is concerned with the association between human-interpreted
tasks and agent actions, predictability is concerned with how predictable the completion of the task is regarding the current action \cite{zhang2017plan}. So, one way to generate explicable behavior is for the AI agent to use plan distance between the expected and agent action \cite{kulkarni2019explicable}.\\ While the works here usually represent the model as state information, goal and initial condition, the human observational model will be added into the model representation if the AI agent tries to express its intentions with legible behavior which enables the human collaborators to infer the goal \cite{dragan2013legibility}. In \cite{dragan2013generating}, they proposed a gradient optimization technique to autonomously generate legible motion, and used a trust level constraint to control the unpredictability of that motion. Transparent planning is also committing in communicating goal, while the AI agent can communicate their goals through action as efficient as possible regardless of how far this might remove it from the goal \cite{macnally2018action}. Even though it is proven that legible and predictable behaviors affect the collaboration fluency \cite{dragan2015effects}, the AI agent should be able to obfuscate the plan to protect privacy in the case of having adversarial entities \cite{kulkarni2019unified}. Thus, it is very important that the AI agent synthesizes a single behavior that is simultaneously legible to friendly entities and obfuscatory to adversarial ones \cite{kulkarni2019signaling}.\\ 
In addition to communicating intentions through behavior, the AI agent can communicate its incapabilities through showing what and why it is unable to accomplish \cite{kwon2018expressing}. The mentioned works in this line mostly consider sequential task horizon, however, there are another set of works that account for the human mental model to improve team performance when there is a single prediction task with error boundary as the model \cite{bansal2020optimizing,bansal2019updates}.
 %\vspace{-6pt}
\subsection{Interpretable Through Explanation}
It is necessary for the AI agent to be able to provide explanations to its human collaborator to increase the interpretability of its behavior. For a single classification task, the explanation can involve explaining the correctness or rationale of the decisions such as providing faithful and customized explanations of a black box classifier that accounts for the fidelity to the original model as well as user interest \cite{lakkaraju2019faithful}, or the explanation can concern with analyzing and explaining the details of failure and error \cite{nushi2018towards}, which is good for debugging and troubleshooting. Moreover, the explanation can be called upon to explain the AI agent's incomprehensible behaviors or plans which are categorized as sequential tasks. This explanation can solve the root cause of inexplicable behavior, in which the AI agent provides explanations to reconcile the human model to its model till the behavior becomes explicable to the human \cite{chakraborti2017plan,sreedharan2017explanations}. However, the comprehensible behavior might be infeasible, in which case the explanation can be in the form of expressing incapability. This can come in the form of explaining the unsynthesizable cores of a specified behavior \cite{raman2013sorry,raman2013towards,briggs2015sorry}, or the absence of a solution to a planning task (unsolvability). Explaining the unsolvability of a planning problem can be in the form of providing a certificate of unsolvability \cite{eriksson2017unsolvability,eriksson2018proof}, or a more compact and understandable reason through the use of hierarchical abstraction to generate a reason for unsolvability \cite{sreedharan2019can}. Furthermore, an AI agent can provide a novel behavior that makes a trade-off between explanation and explicable behavior to combine the strength of each \cite{chakraborti2017balancing,sreedharan2019expectation}.
%\vspace{-9pt} 
\section{Bidirectional Complementing}
Instead of having one of the AI agent or the human be responsible for the task and the other acting as an advisor or observer, here we have cases in which both the AI agent and the human are responsible for the task and each of them helps the other in the task or decision making. So, both the AI agent and the human may alternatively enter the land of one another in performing a task, decision making, or coordination. This is considered as bidirectional complementing because either the communications or the actions are bidirectional. Therefore, we can categorize it into bidirectional communication or behavioral coordination which results in more effective team performance. As a result, the teaming goal for works in this category is all to improve the team performance. Also, most of the works in this category include planning and scheduling tasks which makes the task horizon as sequential tasks.
 %\vspace{-6pt}
\subsection{Bidirectional Communication}
One of the important challenges is \textit{when} and \textit{what} to communicate during human-AI collaboration. For example, \cite{unhelkar2020decision} proposed a CommPlan framework which enables jointly reasoning about the robot's action and communication at its policy in a shared workspace task where the robot has multiple communication options and need to reason in a short time. Also, \cite{grover2020model} investigated robot reasoning on how to interact to localize the human model by generating the right questions to refine the robot understandings of the teammate.\\
Other challenges arise when the AI agent and the human participate in an interactive dialogue such as contrastive explanation in the framework of counterfactual reasoning. In a planning problem \cite{sreedharan2018hierarchical} or a scheduling problem \cite{zahedi2020not}, this contrastive explanation would help the understanding of the user who is confused by the agent's behavior or offers and presents an alternative behavior that they would expect. Also, this can be used for model refinement, like the RADAR-X framework that uses the foil raised by the user as evidence for unspecified user preferences and to refine plan suggestions \cite{valmeekam2020radar}.
 %\vspace{-6pt}
\subsection{Behavioral Coordination}
Behavioral coordination includes both team level and task level coordination. In task-level coordination, an AI agent can coordinate its behavior for a serendipitous plan in a cohabitation scenario \cite{chakraborti2015planning}, or it can employ human motion prediction in conjunction with a complete, time-optimal path planner to execute efficient and safe motion in the shared environment \cite{unhelkar2018human}, and by detecting blind spot of both the human and the AI agent, they can coordinate for safe joint execution by handing off the task to the most capable agent \cite{ramakrishnan2019overcoming}. Also, the AI agent can use nonverbal cues and feedback to signal how it expects the human to act next to enable the human to demonstrate their preferences more effectively \cite{huang2019nonverbal}. Furthermore, in a team level coordination, the AI agent coordinates at a team level. For example, Mobi is a single interface that enables crowd participants to tackle tasks with global constraints. Mobi allows the users to specify their desires and needs, and produce as output an itinerary that satisfies the mission \cite{zhang2012human}. AI-mix is an interface that improves the effectiveness of human crowds. It aims at planning and scheduling tasks for crowds by facilitating roles such as steering and interpretation \cite{manikonda2014ai}. RADAR, a proactive decision-making system, improves the decision-making experience of the human by providing suggestions that aid in constructing a plan for single \cite{sengupta2017radar,grover2020radar} and multiple \cite{sengupta2018ma} humans.
\section{Goal of Teaming}
It is very important to see what the purpose of the human and the AI agent is in forming a team, so depending on the main purpose of teaming, other factors may significantly change. In other words, depending on the performance they seek and the interaction state, the team goal will be different.\\ 
\textbf{Performance goal } Despite the commonly accepted assumption that performance goals of individual entities in the team will result in better overall team performance, it is shown that in some cases better individual performance cannot cause better team performance due to incompatibility between them \cite{bansal2019updates}. Thus, when we have a team of human and AI, if the team is formed to improve the team performance, it is very necessary to take into account the whole team performance instead of individual performance.\\
\textbf{Interaction State } is a significant factor that affects the teaming goal. If the human and the AI form a team for a short or single interaction it will affect the setting and general teaming goal differently than longitudinal interactions. For instance, the interpretability concepts are all meaningful when there is longitudinal interactions between the human and the AI. Besides, other concepts such as trust emerge which might not be important in a single interaction. While in a single interaction teaming, both the human and the AI seek immediate rewards, in longitudinal interactions rewards over a longer horizon are important, which influence the teaming strategy significantly.

%\vspace{-8pt} 

\section{Conclusion}
This survey provides an overview of the many different directions of works in human-AI symbiosis and the current trends in this area. Generally, human-AI symbiosis is a growing field of research with variety of challenging problems. Since the recent works in this area explored the existing challenges and potential solutions from different perspectives, the lack of clear connection makes them seem independent. This issue limits the use and expansion of one method from one direction to another. Therefore, in this paper, we highlighted the various dimensions that ramify the researches in this area and as a result diverge the researchers from finding a clear connection between their works. We emphasized some of the important angles in this area (1) complementing flow (2) task horizon and model representation (3) knowledge and capability level (4) teaming goal. We noted that all the researches in this area are matched into one category of each dimension, then we have a collection of works that are mutual in all the mentioned directions. With the clustering of works based on their common dimensions, not only works of a similar nature are easily identifiable but it also provides the potential for works that fall in different clusters to find common ground. We hope that this survey makes a direction for future research and provide a clear connection between the works in the field such that future connections between methods and solutions can be used from the different dimensions.  %\section{Technique and Model Representation}

\noindent \paragraph{\textbf{Acknowledgments.} } This research is supported in part by ONR grants N00014-16-1-2892, N00014-18-1- 2442, N00014-18-1-2840, N00014-9-1-2119, AFOSR grant FA9550-18-1-0067, DARPA SAIL-ON grant W911NF-19- 2-0006, NASA grant NNX17AD06G, and a JP Morgan AI Faculty Research grant.
\bibliographystyle{plain}
\bibliography{ijcai21}
\end{document}